\pdfoutput=1

\documentclass[11pt]{article}
\usepackage{naacl2021}

\usepackage{times}
\usepackage{latexsym}
\usepackage[algoruled,ruled,vlined,noend]{algorithm2e}
\usepackage{times}
\usepackage{latexsym}
\usepackage{pgfplots}
\usepackage{graphicx}
\usepackage{enumitem}
\usepackage{xcolor,colortbl}
\usepackage{tikz}
\usepackage{diagbox}
\usepackage{tkz-tab}
\usepackage{caption}
\usepackage{latexsym}
\usepackage{amssymb}
\usepackage{amsmath}
\usepackage{subcaption}
\usepackage{natbib}
\usepackage{blindtext}
\usepackage{url}
\usepackage{color}

\usepackage{url}
\usepackage{hyperref}    
\usepackage{cleveref}
\SetAlFnt{\small}
\SetAlCapFnt{\small}
\SetAlCapNameFnt{\small}
\usepackage{ntheorem}
\newtheorem{hyp}{Hypothesis }

\usepackage[T1]{fontenc}
\usepackage[utf8]{inputenc}
\usepackage{microtype}

\title{The Curious Case of Hallucinations in Neural Machine Translation}

\author{Vikas Raunak\qquad Arul Menezes\qquad Marcin Junczys-Dowmunt\\
~Microsoft, USA \\
\texttt{\{viraunak, arulm, marcinjd\}@microsoft.com}}

\begin{document}
\maketitle
\begin{abstract}
In this work, we study hallucinations in Neural Machine Translation (NMT), which lie at an extreme end on the spectrum of NMT pathologies. Firstly, we connect the phenomenon of hallucinations under source perturbation to the Long-Tail theory of \citet{feldman_theory}, and present an empirically validated hypothesis that explains hallucinations under source perturbation. Secondly, we consider hallucinations under corpus-level noise (without any source perturbation) and demonstrate that two prominent types of natural hallucinations (detached and oscillatory outputs) could be generated and explained through specific corpus-level noise patterns.  Finally, we elucidate the phenomenon of hallucination amplification in popular data-generation processes such as Backtranslation and sequence-level Knowledge Distillation. We have released the datasets and code to replicate our results at \url{https://github.com/vyraun/hallucinations}.

\end{abstract}

\section{Introduction}
\label{sec:intro}
Neural Machine Translation (NMT) enjoys tremendous success, far surpassing the performance of previous statistical approaches in high-to-moderate resource settings \cite{koehn}. However, NMT suffers from well known pathologies such as coverage \cite{coverage}, mistranslation of named entities \cite{ugawa}, etc. In terms of adequacy of the generated output \cite{adequacy}, hallucinations are egregious mistakes that lie at the extreme end of NMT pathologies. Such hallucinated outputs are characterized as being decoupled from the source sequence, despite being (fully or moderately) fluent in the target language \cite{muller_definition}. Two main hallucination phenomena have been reported in the existing literature:

\begin{enumerate}
    \item NMT models tend to generate hallucinated outputs under certain cases of source perturbation \cite{lee_hallucinations}.
    \item NMT models have a propensity to hallucinate more frequently under out-of-domain inputs \cite{muller_definition}.
\end{enumerate}

However, a plausible theory to explain the generation of different types of hallucinations, including the above two results is still lacking in the NMT literature. \citet{lee_hallucinations} posited that hallucinations could be happening due to decoder instability, however, their experiments to engineer solutions based on this proved inconclusive. In this work, we present a systematic study of different kinds of hallucinations, studying them through the lens of generalization, memorization and optimization in sequence to sequence models. Our key contributions are as follows:
\begin{enumerate}
    \item We extend the Memorization Value Estimator proposed in \citet{feldman_main} to the sequence to sequence setting and demonstrate that hallucinations under source-side perturbations could be explained through the long-tail theory they propose.
    \item We introduce corpus-level noise into NMT parallel corpora and show that specific noise patterns interact with sequence to sequence training dynamics in different ways to generate the prominent hallucination patterns reported in the literature \cite{lee_hallucinations}.
    \item We demonstrate the phenomenon of hallucination amplification in the outputs generated using Backtranslation \cite{bt_at_scale} and Knowledge Distillation \cite{kd_sequence}, two widely used data generation algorithms for MT.
\end{enumerate}

\section{Related Work}
\label{sec:related_work}
Our work connects hallucinations in NMT to the problem of generalization in Deep Learning. In this section, we briefly survey the two areas.

\subsection{Hallucinations in NMT}
The phenomena of hallucinations in NMT lack clear categorical definitions. \citet{lee_hallucinations} define hallucinations as the model producing a vastly different (inadequate) output when the source is \textit{perturbed} under a specific noise model and present an algorithm to detect such cases. Subsequently, approaches to making NMT models more robust to small perturbations in the input have been actively explored \cite{google_robustness}, however, no coherent theory to explain the phenomena of hallucinations has been empirically validated in the existing literature. Our work differs from \citet{lee_hallucinations} in that we not only study hallucinations under source side perturbations but also under corpus-level noise. Further, we build on their work by filling in the gap for a plausible hypothesis that explains various types of hallucinations.

\citet{mueller_2} consider hallucinations as outputs detached from the source, and demonstrate that NMT models are more prone to hallucinations under out-of-domain settings by manually ascertaining whether an output generated is hallucinated or not. Manual detection of hallucinations, however, is an impediment for fast experimental cycles, and in this work, besides explaining the generation of such \textbf{\textit{natural} hallucinations} (i.e. hallucinations generated without any source perturbation), we also propose an approximate corpus level hallucination detection algorithm to aid faster analysis.

\subsection{Generalization in Deep Learning}
\citet{feldman_theory} studies label memorization in deep learning, and explains how memorization could be essential for achieving close-to-optimal generalization when the data distribution is long-tailed; since memorizing a representative of a rare subpopulation from the long-tail could significantly increase the prediction accuracy on its subpopulation, thereby improving the generalization error. Follow-up work \cite{feldman_main} empirically validates the key ideas of this \textbf{long tail theory} by making use of a memorization estimator to test \textbf{its} predictions for classification problems. To the best of our knowledge, our work presents the first study that connects Feldman's long-tail theory to the problem of hallucinations in NMT.

\section{Categorizing Hallucinations in NMT}
\label{sec:categorization}
In this section we systematize the study of hallucinations by coining a few definitions to aid further analysis. Firstly, we categorize hallucinations in NMT into two primary categories:

\begin{enumerate}
    \item Hallucinations under Perturbations (HP): For a given input source sequence, a model is considered to generate a hallucination under perturbation, if the generated translations for perturbed and unperturbed sequences differ drastically. More precisely, we refer to the algorithm proposed by \citet{lee_hallucinations} for detecting hallucinations under perturbation.
    \item Natural Hallucinations (NH): For a given unperturbed input source sequence, a model is considered to generate a natural hallucination if the generated translation is severely inadequate (fluent or otherwise).
\end{enumerate}

\begin{figure}[ht]
\centering

\fbox{\parbox{\dimexpr\linewidth-2\fboxsep+2.5\fboxrule\relax}{ 
\textbf{Source}: das kann man nur feststellen , wenn die kontrollen mit einer großen intensität durchgeführt werden .  \\
\textbf{Correct Translation}: this can only be detected if controls undertaken are more rigorous . \\
\textbf{Output}: blood alone moves the wheel of history , i say to you and you will understand , it is a privilege to fight .
}}%

\caption[]{Detached Natural Hallucination Example}
\label{fig:dh}
\end{figure}

\begin{figure}[ht]
\centering

\fbox{\parbox{\dimexpr\linewidth-2\fboxsep+2.5\fboxrule\relax}{ 
\textbf{Source}: 1995 das produktionsvolumen von 30 millionen pizzen wird erreicht .  \\ 
\textbf{Correct Translation}: 1995 the production reached 30 million pizzas . \\
\textbf{Output}: {\color{orange} the us} , for example , {\color{red}has been in the} past two decades , but {\color{red}has been in the} same position as {\color{orange}the us} , and {\color{red}has been in the} united states .
}}%

\caption[]{Oscillatory Natural Hallucination Example: Decoupled from Source + Repeating N-gram Structure}%
\label{fig:oh}%
\end{figure}

Further, we classify a Natural Hallucination (NH) as belonging to one of the two types:
\begin{enumerate}
    \item Detached Hallucinations (DH): A fluent but completely inadequate translation (e.g. Figure \ref{fig:dh}).
    \item Oscillatory Hallucinations (OH): An inadequate translation that contains repeating n-grams (e.g. Figure \ref{fig:oh}).
\end{enumerate}

Both Figures \ref{fig:dh} and \ref{fig:oh} show the tokenized input and output (hallucinated) examples from models trained in Section \ref{ssec:natural_hal}, to illustrate the above two definitions. The above categorization of Natural Hallucinations excludes two other types of pathologies, discussed as hallucinations in \citet{lee_hallucinations}, namely, generation of shorter outputs and copy of source to the output. The proposed categorization allows us to quantitatively disentangle the study of hallucinations from other NMT pathologies, without losing any generality.

\section{Origins of Hallucinations}
\label{sec:origins}

In this section, we propose and empirically validate two hypotheses in order to explain the two categories of hallucinations described in section \ref{sec:categorization}.
 
\subsection{Hallucinations under Perturbations}
\label{ssec:hp}

\begin{hyp}[H\ref{hyp:first}]
\label{hyp:first}
The samples memorized by a NMT model are most likely to generate hallucinations when perturbed.
\end{hyp}

To validate H1, we adapt the Memorization Value Estimator (MVE) proposed by \citet{feldman_main} to the sequence to sequence setting, by replacing the accuracy metric they use with a sequence overlap metric such as chrF \cite{popovic-2015-chrf} or BLEU \cite{papineni-etal-2002-bleu}\footnote{In practice, other MT metrics such as METEOR or BERT-Score \cite{meteor, bert_score} could also be used as empirical extensions of MVE for sequences, however, word/character n-gram overlap provides a stronger indication of memorization than soft-overlap methods like BERT-Score.}. We then compare the hallucination behaviour under perturbation of the most-memorized samples with random samples using the hallucination detection algorithm proposed in \citet{lee_hallucinations}.

\paragraph{Memorization Value Estimation} The modified Memorization Value Estimator (MVE) is described in algorithm \ref{algo:feldman}. MVE computes the memorization value of a sample as the change in average prediction metric $M$ (for which we use metrics such as chrF, BLEU) for the given sample between the models trained with the sample included in the training set and the models trained with the sample excluded. 

\begin{algorithm}[ht]
 \SetAlgoNoLine
 \SetNoFillComment
 \KwData{Training Dataset S of size $n$, Learning Algorithm A, Number of Trials t, Metric M}
 \KwResult{Memorization Values over S}
 Sample $t$ random subsets of $S$ of size $m$\;
 \For{k = 1 to t}{
 Train model $h_k$ by running $A$ on $S_k$
 }
\For{i=1 to n}{
 mem(A, S, i) = $\underset{h \leftarrow A(S)}{E[M]} - \underset{h \leftarrow A(S^{\textbackslash i})}{E[M]}$ }
\caption{Memorization Value Estimator \label{algo:feldman}}
\end{algorithm}

\paragraph{Hallucination Detection} The HP detection algorithm used is presented as algorithm \ref{algo:lee_perturb}. In practice, algorithm \ref{algo:lee_perturb} is a specific instance of the algorithm from \citet{lee_hallucinations}, wherein we make the following three changes:
\begin{enumerate}
    \item We perturb word-tokenized sentences, rather than applying perturbations on BPE-tokenized inputs.
    \item We report results for the perturbation (insertion) at the first position only, which, based on the ablation studies in \citet{lee_hallucinations}, is the most reliable way to generate hallucinations.
    \item We sample the set of perturbation tokens T from the most common tokens in the token dictionary computed over the training corpus, for obtaining the most plausible perturbations.
\end{enumerate}

 \begin{algorithm}[h!]
     \SetAlgoNoLine
     \SetNoFillComment
     \KwData{NMT Model, Parallel Corpus (X, Y), Token Set T}
     \KwResult{Hallucinated Samples $H$}
     \For{x, y in X, Y}{
     $y'$ = Model($x$) \\
     \If {adjusted-bleu($y'$, $y$) > 0.09}
     {\For{$t$ in T}{$\widetilde{x}$ = put $t$ at the beginning of the input $x$ \\
        $\widetilde{y}$ = Model($\widetilde{x}$) \\
        \If{adjusted-bleu($\widetilde{y}$, $y'$) < 0.01}{add x to $H$}
     }
     }
     }
     
 \caption{Hallucination under Perturbation}
 \label{algo:lee_perturb}
 \end{algorithm}

\subsubsection{Experiments and Results}
To compute the memorization values, \textit{mem} in algorithm \ref{algo:feldman}, we train $t=10$ NMT models using fairseq \cite{fairseq} on different randomly selected subsets of sentence pairs (each about $101K$ samples) from the IWSLT-2014 De-En dataset ($160K$ samples). BPE \cite{bpe} with a joint token vocabulary of 10K is applied over lowercased tokenized text. The NMT model is a six-layer Transformer model with embedding size 512, FFN layer dimension 1024 and 4 attention heads (42M parameters), and the checkpoint with the best validation BLEU (detokenized, with beam=5) is selected. In each case, a batch size of 4K tokens, dropout of 0.3 and tied encoder-decoder embeddings is used. Then, the MVE (algorithm \ref{algo:feldman}) is applied on the training samples using the above $t$ trained models to compute the memorization values, \textit{mem} for each source sample $i$. For further analysis, we do not consider any sample which hasn't been excluded from the random training sets at least twice.

To generate HP we use algorithm \ref{algo:lee_perturb} with the set $T$ consisting of 30 tokens randomly sampled from the top 100 most common tokens. We apply algorithm \ref{algo:lee_perturb} to two sets of training samples -- a Memorized set comprising of training samples with the highest hundred (100) memorization values, and a Random set (of the same size) sampled from the rest of the training samples. Since, each input sentence can appear in the Hallucinated Samples set $H$ multiple times in algorithm \ref{algo:lee_perturb}, we report both Unique and Total number of Hallucinations (HP) generated.

We report results using chrF, BLEU as well as the prediction accuracy computed by matching the entire output string to the reference, as the metric $M$ used in computing the memorization values. Table \ref{tab:chrf} shows that the difference between the counts of unique HP between the Memorized and Random set is very high. The same trend holds using BLEU and prediction accuracy as metrics as well (Tables \ref{tab:bleu}, \ref{tab:pacc}), even though as the metric for computing memorization values becomes more coarse-grained (going from chrF to accuracy), the differences get reduced.

\begin{table}[!htbp]
\centering
\scalebox{0.9}{
\begin{tabular}{c|c|c}
\hline
Set & \textbf{Unique HP} & \textbf{Total HP} \\ \hline
\textbf{Random}  & 1 & 1 \\ 
\textbf{Memorized}  &  \cellcolor{orange!50} 51 & \cellcolor{red!65} \textbf{431} \\ \hline
\end{tabular}}
\caption{Memorized vs Random Set Comparison using Algorithm \ref{algo:lee_perturb} with chrF as the Metric in Algorithm \ref{algo:feldman}.}
\label{tab:chrf}
\vspace{-0.5em}
\end{table}

\begin{table}[!htbp]
\centering
\scalebox{0.9}{
\begin{tabular}{c|c|c}
\hline
Set & \textbf{Unique HP} & \textbf{Total HP} \\ \hline
\textbf{Random}  & 2 & 2 \\ 
\textbf{Memorized}  &  \cellcolor{orange!50} 42 & \cellcolor{red!65} \textbf{180} \\ \hline
\end{tabular}}
\caption{Memorized vs Random Set Comparison using Algorithm \ref{algo:lee_perturb} with BLEU as the Metric in Algorithm \ref{algo:feldman}.}
\label{tab:bleu}
\vspace{-0.5em}
\end{table}

\begin{table}[!htbp]
\centering
\scalebox{0.9}{
\begin{tabular}{c|c|c}
\hline
Set & \textbf{Unique HP} & \textbf{Total HP} \\ \hline
\textbf{Random}   & 4 & 5  \\
\textbf{Memorized}   & \cellcolor{orange!50} 8 & \cellcolor{orange!50} 30  \\ \hline
\end{tabular}}
\caption{Memorized vs Random Sets Comparison using using Algorithm \ref{algo:lee_perturb} with Accuracy as the Metric in Algorithm \ref{algo:feldman}.}
\label{tab:pacc}
\vspace{-0.5em}
\end{table}

\begin{figure}
\centering
\begin{subfigure}[b]{0.45\textwidth}
\centering
% \captionsetup{font=small, skip=0pt}
\includegraphics[height=5.0cm, width=\textwidth]{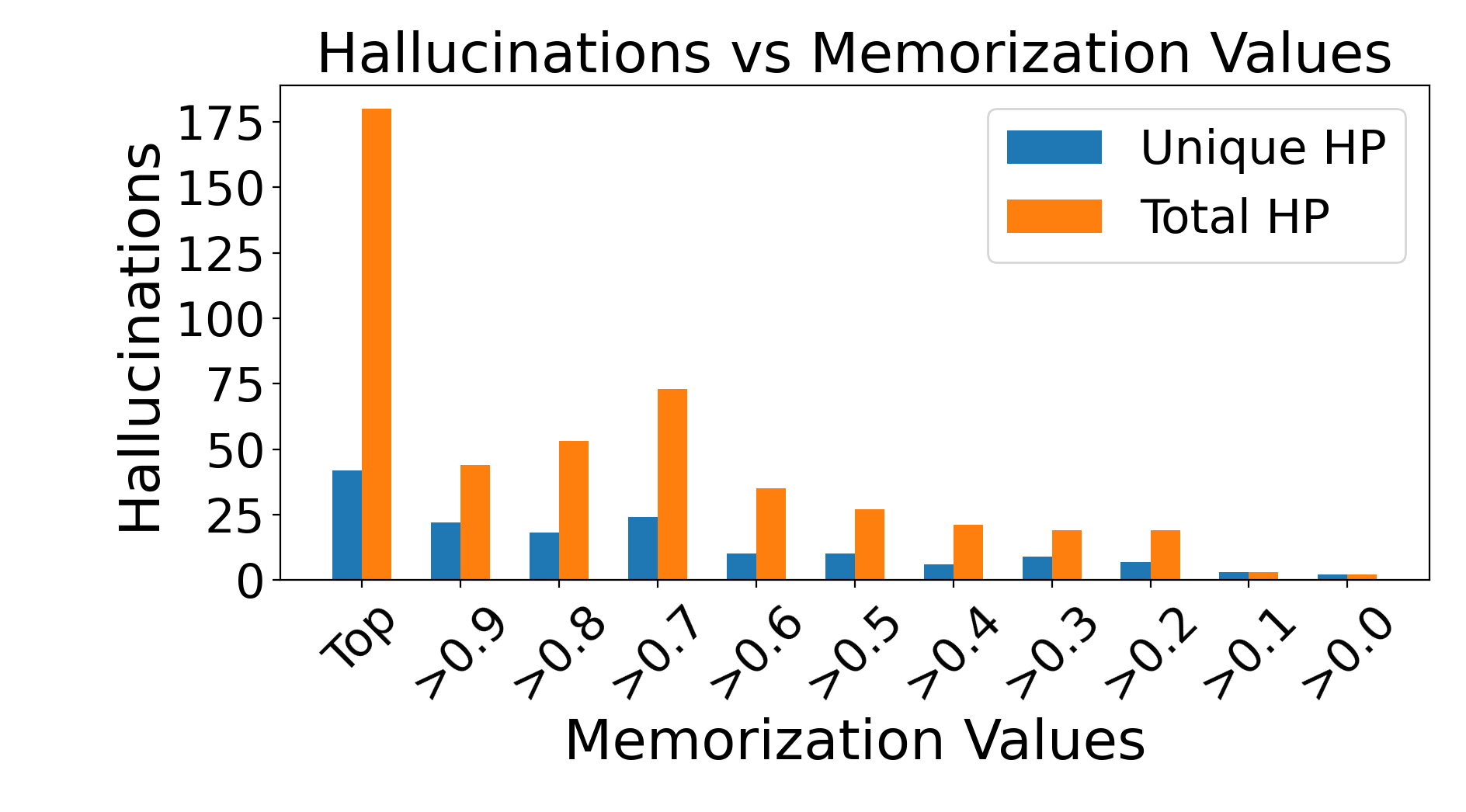}
\end{subfigure}
\begin{subfigure}[b]{0.45\textwidth}
\centering
\captionsetup{font=small, skip=0pt}
\includegraphics[height=5.0cm, width=\textwidth]{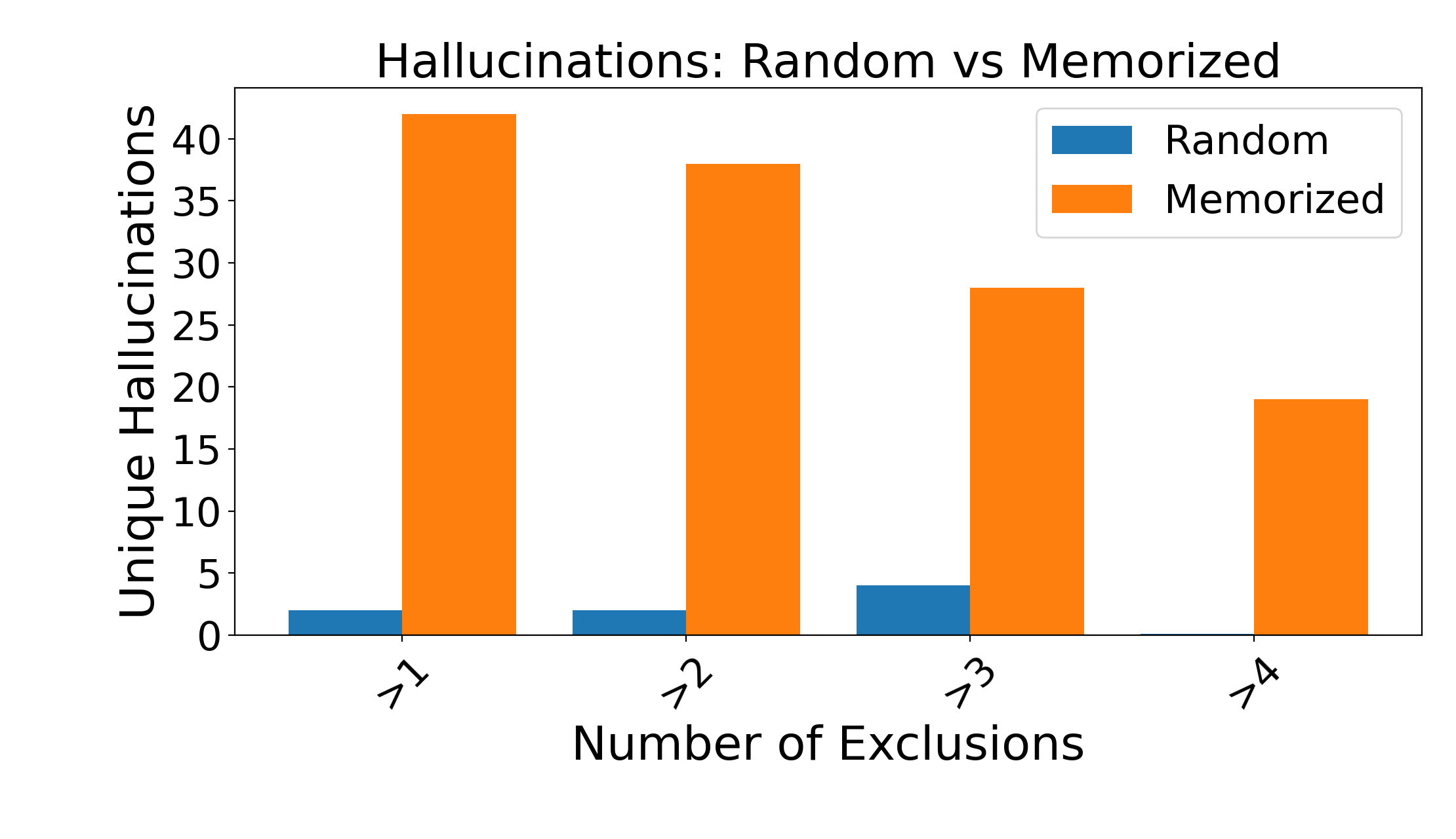}
% \caption{TreeDepth}
% \label{fig:ling_prob_tree}
\end{subfigure}
\vspace{-0.25em}
\caption{Further Comparisons: (Top) measures Hallucinations under increasingly restrictive sampling sets, in terms of the memorization value. (Bottom) compares the Memorized vs Random sets under different number of sample exclusions}
\label{fig:full_an}
\vspace{-1.25em}
\end{figure}

\paragraph{Further Comparisons}Figure \ref{fig:full_an} (Top) presents the number of unique hallucinations (using BLEU as the metric in algorithm \ref{algo:feldman}, as in Table \ref{tab:bleu}; the default metric from hereon, unless stated otherwise), when the underlying sampling set for constructing the set under evaluation, is restricted using different threshold memorization values (varying from 0 to 0.9, in increments of 0.1). The figure shows that as the memorization values increase, the number of unique (Unique HP) as well as total hallucinations (Total HP) keeps increasing as well, demonstrating a strong positive correlation between hallucination frequency and memorization values. 

Figure \ref{fig:full_an} (Bottom) presents the results for the experiment wherein we refine the memorization value estimates by restricting the Memorized vs Random set comparisons to only the cases when a particular sample has been excluded more than $n$ times (X-axis values) when training the $t$ NMT models. Here, we find that the trend of large differences between the counts of unique hallucinations generated for the two sets stays consistent as the memorization value estimates are made more accurate. In fact, when the two sets (Random, Memorized) are constructed only over the samples which have been excluded at least 4 times, we find zero unique HP for the Random set.

\paragraph{Encoder-Decoder Attention Analysis} To further analyze how memorized samples suffer more hallucinations under perturbations, we compare the cross-attention heads of the last layer of the decoder for the Random and Memorized sets. Table \ref{tab:attn_comp} presents a comparison of the average entropy of the attention matrix, averaged diagonal attention and the average attention paid to the last source token, aggregated over the entire sets. The results show that the two sets differ considerably in terms of the attention distribution, with the memorized set having more fixed (lower-entropy) average attention distributions. Although this result is known for hallucinated translations \cite{lee_hallucinations, voita2020analyzing, berard-etal-2019-naver}, which have a tendency of producing deficient attention maps, the fact that this phenomenon extends to memorized samples as well further helps establish the link between memorization and hallucination under perturbation.

\begin{table}[!htbp]
\centering
% \captionsetup{font=small}
%\vspace{-0.8em}
\scalebox{0.8}{
\begin{tabular}{c|c|c}
\hline
% \backslashbox{\textbf{Embeding}}{\textbf{LangPair}}   
Data-Type & \textbf{Memorized} & \textbf{Random} \\ \hline
Attention Entropy   & \cellcolor{yellow!50}\textbf{1.85}  & \cellcolor{red!65}\textbf{2.70} \\ 
Diagonal Attention Entropy   & \cellcolor{yellow!50}1.48  & \cellcolor{red!65}2.21 \\ \hline
Average Last Token Attention   & \cellcolor{yellow!50}\textbf{0.35}  & \cellcolor{red!65}\textbf{0.25} \\ \hline
\end{tabular}}
\caption{Attention Statistics Comparison for Random vs Memorized Sets.}
\label{tab:attn_comp}
\vspace{-0.5em}
\end{table}

\subsection{Natural Hallucinations}
\label{ssec:natural_hal}

\begin{hyp}[H\ref{hyp:two}] \label{hyp:two}
Corpus-level noise patterns (comprised of invalid source-target pairs) dictate the type of natural hallucinations generated by the NMT model. 
\end{hyp}

Hypothesis 2 posits the simplest explanation for the generation of natural hallucinations: that the phenomenon is caused by the presence of invalid references in the training data, and that \textit{specific} patterns of such corpus-level noise cause \textit{specific} hallucination patterns to emerge. Establishing a causal link between corpus-level noise patterns and hallucination types could greatly ease diagnosing the origins of such cases. 

We try to validate H2 by construction: first, we build four different types of the corpus-level noise patterns, and then we analyze the resulting models in terms of the generated translations.

\subsubsection{Experiments and Results}

We train 5 models on the IWSLT 2014 corpus, where the training data consists of 160K samples. We train a baseline model with no noise, while the other 4 models are trained with specific patterns of added noise. The model and training settings are the same as in section \ref{ssec:hp}, except that BPE is now learnt on the noise-added corpus for the 4 models.

\paragraph{Corpus-Level Noise Model} In order to generate the noise sets to be added to the training parallel data, we first construct an invalid reference set (IRS), a small set of detached source-target pairs and use the larger WMT 2014 De-En corpus as an additional data source (the size of the constructed IRS is 21 for the below experiments). Then, the different noise sets (of the same size) are constructed using different sampling strategies for sources and targets, which combine source-target sequences drawn from the IRS and the WMT 2014 De--En training corpus into noise sets with particular characteristics. Specifically, we generate the noise sets as follows:

\begin{enumerate}
    \item Unique-Unique (UU): We sample 21K \footnote{21K amounts to approximately 12\% noisy samples, when combined with the 160K parallel training samples for the IWSLT De-En corpus.} random unique source sentences from WMT, and pair each with an unrelated unique random target sentence from WMT.
    \item Repeat-Repeat (RR): We sample 21 unique source sentences from IRS, and pair each with unrelated unique random target sentence from IRS, and repeat each such pair 1000 times.
    \item Repeat-Unique (RU): We use the same 21 random unique source sentences as RR. We repeat each 1000 times, and pair each repeat with unrelated unique random target sentence from WMT.
    \item Unique-Repeat (UR): We sample 21 random unique target sentences from the IRS. Each such target sentence is repeated 1000 times. Each repeat is paired with an unrelated unique random source sentence from WMT.
\end{enumerate}

\paragraph{Evaluation} We train NMT models with each of the above four noise sets added to the IWSLT De-En parallel corpus, and report the results for both De-En and En-De translation directions. Specifically, we investigate the behavior of models trained on each of the above noise sets using the following evaluation sets: 

\begin{enumerate}
    \item IWSLT: The IWSLT De-En 2014 test set, which does not overlap with the training data, is used to measure generalization.
    \item Invalid reference set (IRS): The 21 unique source-target sentence pairs in the IRS are also used as an evaluation set. Due to the way the noise sets are built, the IRS overlaps with the various training sets: it is contained in the RR training data, its source sentences are present in the RU training data and its target sentences are present in the UR training data, while there is no overlap for the UU training data. The main purpose of evaluating models on this set is to measure memorization of the overlapping source/targets.
    \item Valid reference set (VRS): This set contains the same 21 source sentences as the IRS, however, they are paired with their valid (correct) references. The VRS set is used to measure whether the NMT model can generalize despite the presence of source/targets associated with the noise sets.
\end{enumerate}

Using the above evaluation sets, we then compute the following metrics:

\begin{itemize}
    \item BLEU: BLEU score for each evaluation set. 
    \item IRS-NH: We compute the percentage of natural hallucinations (NH) (manually identified) in the translations of the IRS.
    \item IRS-OH: We compute the percentage of oscillatory hallucinations (OH) (manually identified) in the translations of the IRS.
    \item IRS-Repeats: We compute the percentage of the hallucinations that exactly match a reference in the training data.
    \item IRS-Unique Bigrams: We compute the number of unique bigrams in the translations of the IRS, as a fraction of total possible unique bigrams in sentences of the same length.
\end{itemize}

\begin{figure}
\label{fig:main_sample}
\centering%

\fbox{\parbox{\dimexpr\linewidth-2\fboxsep+2.5\fboxrule\relax}{\textbf{Source}: das ist eine unerfreuliche situation , die wir künftig vermeiden wollen . \\ 
\textbf{VRS Reference}: that is an undesirable situation , we do not want that situation in the future .\\
\textbf{No Noise Output}: this is an unpleasant situation that we \&apos;re trying to avoid in the future . \\
\textit{\textbf{UU Output}: the us , in particular , is not alone . \\
\textbf{UR Output}: the football player said that he had never experienced a victory like this . \\
\textbf{RU Output}: the us , for example , has been in the past two decades , but the world has been in the past .\\
\textbf{RR Output}:  that is what she said .} %%% 
}}%
\caption[]{Sample Outputs under Corpus-level Noise}%
\label{fig:main_example}%
\end{figure}

\paragraph{Design of Noise patterns} While the above noise patterns are quite plausible in a web-based corpus collection process, due to the widespread adoption of automatic bitext mining algorithms \cite{bitext_mining} applied over noisy sources, our primary motivation behind constructing these four types of noise patterns is to present different optimization scenarios for the NMT model under training. In each of the four noise patterns, the source-target pairs are `invalid', but the difference lies in the number of representation pathways (contexts) each set offers for the `invalid error' to propagate to the different layers, imposing a different set of requirements on the underlying optimization process. We posit that the four different noise patterns (RU, UR, UU, RR) interact in different ways with the encoder and decoder of an NMT model, e.g. for the RU noise pattern, the decoder is required to generate unique translations for the same sources, thereby encouraging decoder instability, whereas under the UR noise pattern, the encoder is required to produce the same representations for unique inputs, allowing the `invalid error' to propagate to lower encoder layers. In UU noise as well, the model is required to produce encoder representations that are vastly different in the representation similarity space (when compared to the rest of the training corpus), while offering multiple contexts for the invalid error to propagate, while in the case of RR noise, the invalid error propagation is quite restricted. Further, we can test whether the above hypotheses have any predictive power through the properties of the generated translations of noisily trained models. However, a rigorous exploration of the impact of noise patterns on encoder-decoder training dynamics is out of scope for this work.

\begin{table*}
%\vspace{-0.3em}
\centering
%\vspace{-0.6em}
\scalebox{0.90}{
\begin{tabular}{c|c|c|c|c|c|c|c}
\hline
% \backslashbox{\textbf{Embedding}}{\textbf{Tasks}}
Noise & \textbf{IRS-\small{BLEU}}    & \textbf{VRS-\small{BLEU}}    & \textbf{Test-\small{BLEU}}  & \textbf{IRS NH} & \textbf{IRS OH} & \textbf{IRS Repeats} & \textbf{IRS \small{Unique-Bigrams}}\\ \hline
\textbf{U-U}    & 0.48  & 12.90 & 33.75 & 33.33 \% & 0.0 \% & 0.0 \% & 0.80\\ 
\textbf{U-R}    & 12.20 & 12.82 & 33.26 & 42.85 \% & 4.76 \% & 38.09 \% & 0.84\\
\textbf{R-U}    & 0.42 & 0.55 & 34.03 & - & 76.19 \% & 0.0 \% & 0.19\\
\textbf{R-R}    & 100.00 & 0.39 & 33.89 & - & - & - & 0.96\\ \hline
\textbf{None}    & 0.54 & 20.59 & 33.65 & 0.0 \% & 0.0 \% & 0.0\% & 0.925 \\ \hline 
\end{tabular}}
\caption{Analysis of Models trained using different Corpus-level Noise Patterns: De-En}
\label{tab:main_1}
%\vspace{-1.00em}
\end{table*}

\begin{table*}
%\vspace{-0.3em}
\centering
%\vspace{-0.6em}
\scalebox{0.90}{
\begin{tabular}{c|c|c|c|c|c|c|c}
\hline
% \backslashbox{\textbf{Embedding}}{\textbf{Tasks}}
Noise & \textbf{IRS-\small{BLEU}}    & \textbf{VRS--\small{BLEU}}    & \textbf{Test--\small{BLEU}}  & \textbf{IRS NH} & \textbf{IRS OH} & \textbf{IRS Repeats} & \textbf{IRS \small{Unique-Bigrams}}\\ \hline
\textbf{U-U}    & 0.25  & 13.26 & 28.61 & 14.28 \% & 0.0 \% & 0.0 \% & 0.91\\ % 3 natural 
\textbf{U-R}    & 0.42 & 12.15 & 27.93 & 9.52 \% & 0.0 \% & 9.52 \% & 0.95\\ % done 2 natural
\textbf{R-U}    & 0.35 & 0.56 & 28.67 & - & 47.61 \% & 0.0 \% & 0.49\\
\textbf{R-R}    & 100.00 & 0.50 & 28.54 &  & - & - & 0.99\\ \hline
\textbf{None}    & 2.06 & 14.62 & 28.33 & 0.0 \% & 0.0 \% & 0.0\% & 0.95 \\ \hline
\end{tabular}}
\caption{Analysis of Models trained using different Corpus-level Noise Patterns: En-De}
\label{tab:main_2}
\vspace{-1.00em}
\end{table*}

\begin{figure}[ht]
\centering
\begin{subfigure}[b]{0.48\textwidth}
\centering
% \captionsetup{font=small, skip=0pt}
\includegraphics[width=\textwidth]{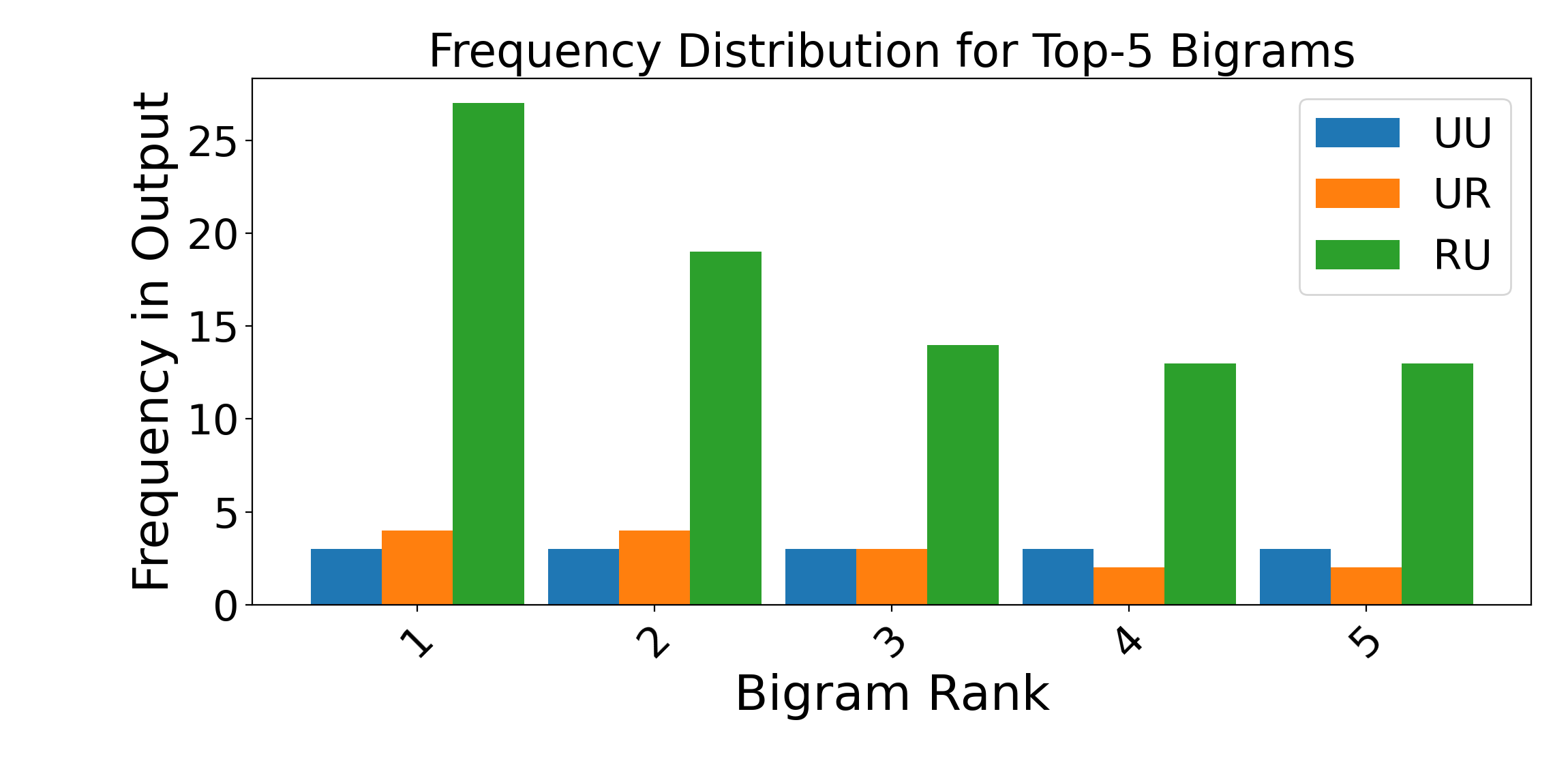} % height=5.0cm, 
\end{subfigure}
\caption{Frequency for the Top 5 Bigrams in the Output for the IRS for the different noisy models on IWSLT 2014 En-De (Table \ref{tab:main_2}).}
\label{fig:bigram}
\end{figure}

\begin{figure}[ht!]
    \centering
    \vskip -3pt
    \includegraphics[width=0.24\textwidth,height=0.25\textwidth]{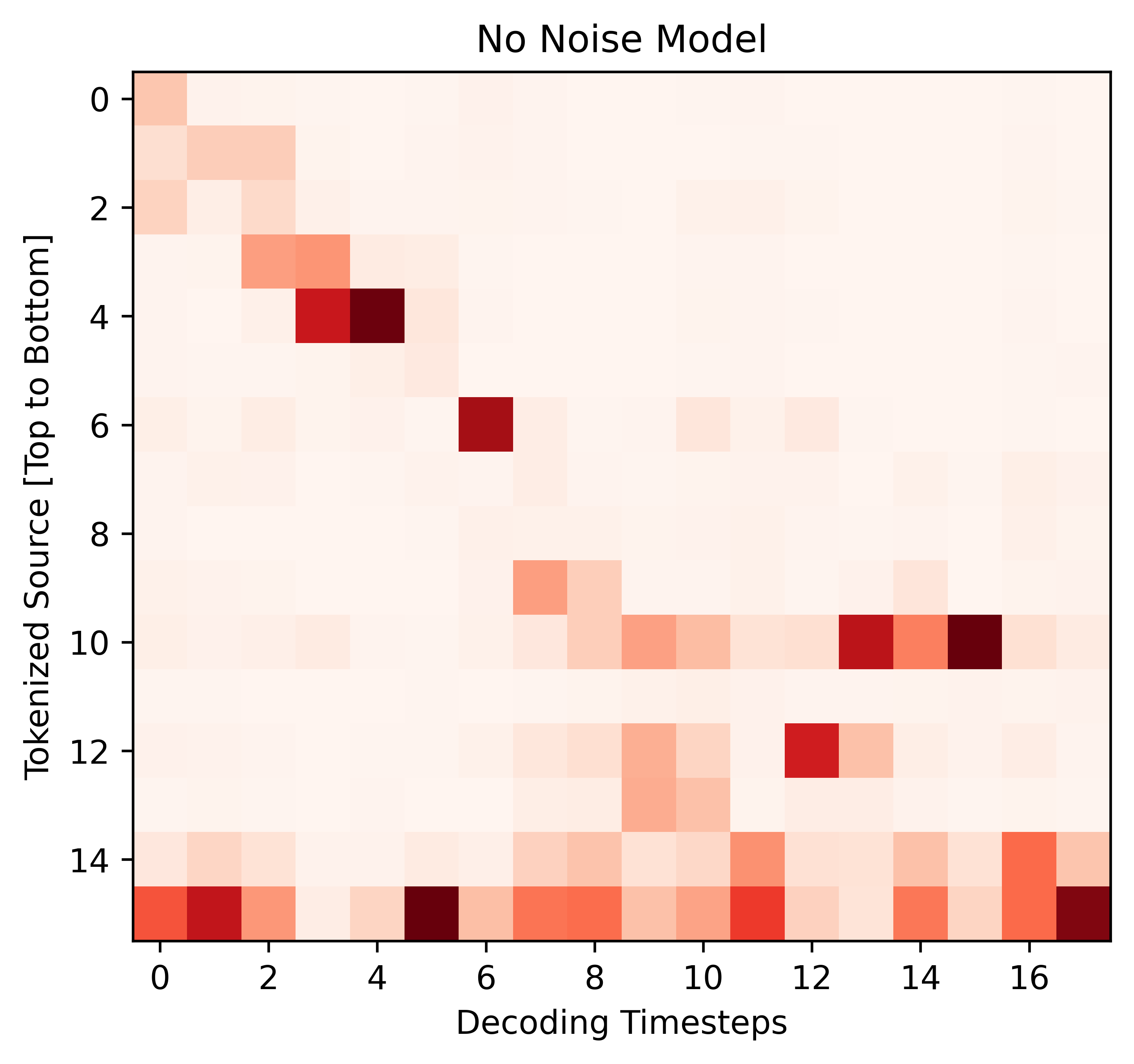}
    \subfloat{\includegraphics[width=0.23\textwidth,height=0.25\textwidth]{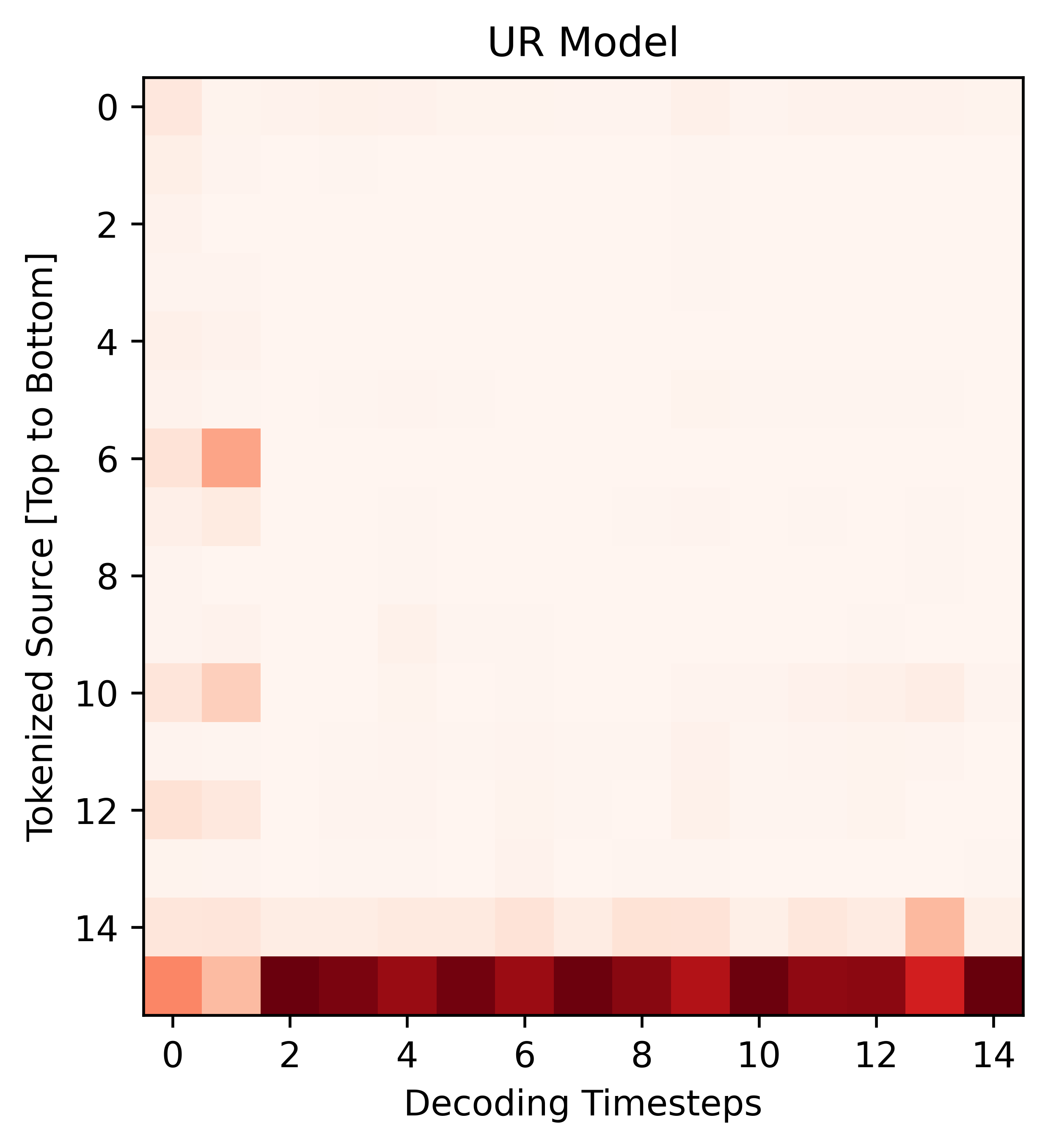}}
    \vskip -7pt
    \caption{Attention Visualization for translation of source in Figure \ref{fig:main_example}, on which the UR model hallucinates (right), compared against the model with no noise (left). The right attention map displays the characteristic hallucination pattern \cite{lee_hallucinations, berard-etal-2019-naver, voita2020analyzing}. The source is not present in the training corpus for the UR model. However, source sequences from the same domain (WMT corpus) are present with invalid references.}
    \label{fig:attention1}
\end{figure}

\paragraph{Results} Tables \ref{tab:main_1} and \ref{tab:main_2} show the results for both the De-En and the En-De translation directions. The boxes marked with `-' are the cases where the associated metric computation does not convey any useful information. We see the following patterns in the results:

\begin{enumerate}
    \item The Test-BLEU is not greatly affected by the noise, except in the UR case, with the models matching the baseline (trained with no noise).
    \item When we consider the IRS-BLEU, we find that the RR model has fully memorized this data. This is to be expected as it has seen this set repeated 1000 times. 
    \item On the IRS set, the UR model produces a number of repeated outputs (IRS Repeats) from the training corpus. 
    \item On the IRS set, the RU model produces a very high percentage of oscillatory hallucinations (OH).
\end{enumerate}

\paragraph{Linking Hallucination Patterns to Noise Patterns} The main purpose of the above experiments is to demonstrate how natural hallucinations can be generated on source sequences seen or unseen during training, and their relation to specific noise types. The link between noise patterns and specific types of hallucinations in the output could be used as very effective diagnostic tool to trace hallucinated outputs to corpus-level noise, with the goal of removing the noise from the training dataset.

In this regard, two important observations further emerge from Tables \ref{tab:main_1} and \ref{tab:main_2}. First, that in the case of UR noise, a considerable percentage of natural hallucinations (IRS NH) manifests as a direct copy of a training reference (without any of the IRS source sequences being present in the training set). Second, for the case of RU noise, oscillatory hallucinations (OH) are very prominent, as evident by the number IRS Unique-Bigrams, which are considerably lower when compared to the other noise types. Figure \ref{fig:bigram} presents the comparisons for counts of the top 5 bigrams present in the translations of the IRS set, showing how among the 4 noise patterns, RU leads to the most oscillatory hallucinations. Resulting sets of translations for a source sequence present in the IRS is shown in Figure \ref{fig:main_example}, while Figure \ref{fig:attention1} presents a qualitative comparison of the attention patterns for this source sequence.

\section{Hallucination Amplification}
\label{sec:hal_amp}
In this section, we analyze how hallucinations caused due to corpus-level noise get amplified when a model trained on a noisy MT corpus is used for downstream data generation in algorithms such as Sequence-level Knowledge Distillation (KD) \cite{kd_sequence} and Backtranslation (BT) \cite{bt_at_scale}. To analyze this, we need to compute NH at scale. So, firstly, we propose an automatic NH detection algorithm based on the analysis that hallucinations often occur in terms of oscillations or repeats of the target sequences.

 \begin{algorithm}[ht]
     \SetAlgoNoLine
     \SetNoFillComment
     \KwData{Source S, Multi-lingual Similarity Scoring Model X, NMT Model M, Noise Estimate $\epsilon$, N-gram order $n$, Threshold $t$}
     \KwResult{Approximate Natural Hallucinations ANH}
     $T$ = Decode the Source Sequences in $S$\;
     $S_{X}$ = Compute Cross-Lingual Similarity for $(T, S)$ \;
     $F_{1}$ = Select Translations where the count of the top repeated $n$-gram in the translation is greater than the count of top repeated source $n$-gram by at least $t$ \;
     $F_{2}$ = Select Translations in $T$ that are Paired with Multiple Unique Sources \;
     $S_{\epsilon}$ = Bottom $\epsilon$ percentage of samples in $S_{X}$ \;
     $ANH$ = $(S_{\epsilon} \cap  F_{1}) \cup  (S_{\epsilon} \cap  F_{2})$
 \caption{Corpus-level NH Estimator}
 \label{algo:corpus_nh}
 \end{algorithm}

The proposed NH Estimator (algorithm \ref{algo:corpus_nh}) is reference-free and works at the corpus-level. One simplifying assumption used in algorithm \ref{algo:corpus_nh} is that the repeats are now computed on the translations generated over the source set rather than on the training set (as in Tables \ref{tab:main_1} and \ref{tab:main_2} for the IRS-Repeats metric). The motivation behind this assumption is that given a sufficiently large source set, the translated output (if hallucinated as a direct copy of one of the training set targets), will appear more than once in the decoded set (since UR noise is one of its causes).  

\subsection{Experiments and Results}
We use algorithm \ref{algo:corpus_nh} to measure NH caused by using the models trained on the noisy corpora (as explored in section \ref{ssec:natural_hal} and analyzed in Tables \ref{tab:main_1} and \ref{tab:main_2}) for BT and Sequence-level KD. For BT, we use 1 million English sentences from the WMT 17 De-En dataset as the monolingual corpus and generate back-translations via sampling \cite{bt_at_scale}, using the different types of noisily trained models (RR, UU, UR, RU) for En-De. For constructing a sequence-level KD dataset we generate the translations over the initial IWSLT 2014 De-En corpus training corpus (the initial parallel data, with no noise) with a beam size of 5 \cite{kd_sequence}. The results of applying the NH estimator (with $\epsilon=1$, i.e. 1\%, $n=4$, $t=2$ and LASER as the cross-lingual similarity scoring Model $M$ \cite{laser}) on the outputs generated using KD and BT are presented in Table \ref{tab:kd} and Table \ref{tab:bt} respectively.

\begin{table}[!htbp]
\centering
\scalebox{0.8}{
\begin{tabular}{c|c|c|c|c|c}
\hline
Data-Type & \textbf{$F_{1}$} & \textbf{$F_{2}$}  & $S_{\epsilon} \cap  F_{1}$ &  $S_{\epsilon} \cap  F_{2}$ &\textbf{$ANH$} \\ \hline
Parallel   & 17 & 1900  & 0  & 41  & 41  \\  \hline
KD-None   & 17  &  2310  & 0  & 120 & 120  \\ \hline 
KD-UU   &  17 &  2394  & 1  & 118 & 119 \\
KD-UR   & 30 & \textbf{3090} & 0 & 688 & 688 \\
KD-RU   &  24 &  2365  & 2 & 122 & 124 \\
KD-RR   & \textit{29}  &  \textit{2411}  & \textit{1 } & \textit{135} & \textit{136}  \\ \hline
\end{tabular}}
\caption{Hallucination Amplification for Knowledge Distillation \cite{kd_sequence}}
\label{tab:kd}
\vspace{-0.5em}
\end{table}

\begin{table}[!htbp]
\centering
\scalebox{0.8}{
\begin{tabular}{c|c|c|c|c|c}
\hline
Data-Type & \textbf{$F_{1}$} & \textbf{$F_{2}$}  & $S_{\epsilon} \cap  F_{1}$ &  $S_{\epsilon} \cap  F_{2}$ &\textbf{$ANH$} \\ \hline
BT-UU   & 154 & 7592 & 0 & 5009 & 5009  \\  
BT-UR   & 43  & \textbf{176184} & 0 & 7280 & 7280  \\  
BT-RU   & 56 & 67 & 1 & 2 & 3 \\  
BT-RR   & \textit{62 } & \textit{65} & \textit{1} & \textit{11 } & \textit{12 } \\ \hline
\end{tabular}}
\caption{Hallucination Amplification for Backtranslation \cite{bt_at_scale}}
\label{tab:bt}
\vspace{-0.5em}
\end{table}

We find that the UR models lead to severe amplifications for both BT and KD. For KD, we find that all noisy models lead to increase in NH when compared to the initial parallel corpus (implying amplification), which itself contains a non-trivial number of repeated targets. For BT, both UU and UR models lead to large number of repeated generations. RR models however cause the least hallucinations for both KD and BT. Our proposed NH estimator is not able to detect many OH however, in any of the cases due to very little overlap with the bottom $\epsilon = 1 \%$ similarity scores, even though the $F_{1}$ column indicates amplification of translations with repeated n-gram patterns ($F_{1}$ ) in the KD datasets.

Further, since, there is hallucination amplification going from a parallel corpus to the KD data generated (using noisy models trained on the parallel corpus), downstream systems trained on the KD data will be impacted in terms of hallucinations as well. We leave further downstream analysis to future work.

\begin{table*}
%\vspace{-0.3em}
\centering
%\vspace{-0.6em}
\scalebox{0.75}{
\begin{tabular}{l|l|l}
\hline
% \backslashbox{\textbf{Embedding}}{\textbf{Tasks}}
MV & \textbf{Source}& \textbf{Target}\\ \hline
0.61    & gerade plus gerade : gerade . ungerade plus ungerade : gerade . & even plus even gives you even . odd plus odd gives you even .\\ % done 2 natural
0.65    & also ist 2 $\wedge$ 5 : 2 x 2 = 4 , 8 , 16 , 32 . & so 2 $\wedge$ 5 is 2 x 2 = 4 , 8 , 16 , 32 . \\ % 3 natural 
0.69    & beweg dich ! nein ! beweg dich ! nein ! beweg dich ! nein ! & move . no . move . no . move . no .\\ % 3 natural 
0.82    & frau : sie bestanden darauf , ich würde lügen . & they insisted that i was lying .\\ 
% 0.80    & flapp ! flapp ! flapp ! & phwap ! phwap ! phwap !\\ % done 2 natural
0.94    & mjam , mjam , mjam , mjam , mjam . & gobble , gobble , gobble , gobble , gobble .\\ \hline
\end{tabular}}
\caption{Examples of Samples from the Top-100 Most Memorized Samples in the Training Set as measured using the Memorization Value (MV) Estimator (Algorithm \ref{algo:feldman}) with chrF as the Metric: De-En.}
\label{tab:mem_examples}
\vspace{-1.00em}
\end{table*}

\section{Discussion}
\label{sec:discussion}
In this section, we present a qualitative analysis of a few topics discussed in section \ref{sec:origins}, along with a discussion on some future research directions.

\subsection{Memorized Samples}
\label{ssec:mem_samples}
Table \ref{tab:mem_examples} presents some examples from the most memorized training samples, thereby representing the samples from the long-tail of the data that is likely to have been memorized by the model. Qualitatively, the examples appear to be different (in terms of source/target syntax) from a random subset of training samples (e.g. in Appendix \ref{sec:appendix}, Table \ref{tab:random_examples}), although we leave further quantitative analysis of the differences to future work. Similarly, the link between out-of-domain and memorized samples needs to be ascertained quantitatively.

\subsection{Preventing Hallucinations}
In this subsection, we discuss a few methods that could be effective in preventing hallucinations.

\paragraph{Data-Augmentation} To prevent hallucinations under perturbation resulting from memorization of the samples in the long-tail of the dataset \cite{feldman_theory}, a simple iterative solution could be to analyze the long-tail (using Algorithm \ref{algo:feldman}), and implement data-augmentations specific to the characteristics of such samples (e.g. as in Table \ref{tab:mem_examples}), with the goal of bringing such samples out of the long-tail \cite{raunak-etal-2020-long}. Further work is required to determine the dynamics of such transition.

\paragraph{Ameliorating Memorization During Learning} Robust learning algorithms e.g. Robust Early learning \cite{early} that are designed to prevent memorization specifically are likely to prevent perturbation based hallucinations.

\paragraph{Robust Learning on Noisy Samples} \citet{truncation} propose a loss-truncation approach to reduce the impact of noisy references in sequence-to-sequence training, using the intermediate model's loss as a sample quality estimator and test their algorithm on a summarization task. \citet{tilted} present a modification to Expected Risk Minimization (ERM), namely Tilted-ERM to reduce the impact of outliers during training. Such techniques could be useful in increasing learning robustness to corpus-level noise in NMT as well.

\paragraph{Corpus-Level Filtering} Incorporating heuristics or filters \cite{dualce, zhang-etal-2020-parallel-corpus} to remove invalid source-target pairs, especially the noise patterns explored in section \ref{ssec:natural_hal} (or to remove bitext indeterminacy in general) could be effective in reducing natural hallucinations.

\section{Conclusion}

In this work we demonstrated that memorized training samples are far more likely to hallucinate under perturbation than non-memorized samples, under an extension of the Memory Value Estimator proposed in \citet{feldman_main}. We also showed that specific noise patterns in the training corpora lead to specific well-known hallucination patterns. Finally, we demonstrated that these patterns can be amplified by popular data-generation processes such as backtranslation and sequence-level knowledge distillation.

Due to the compute-intensive algorithms involved in our analysis, we conduct most of our experiments using the IWSLT 2014 corpus. However, long-tailed phenomena are a characteristic of natural language and even scaling the size of the corpus doesn’t alleviate the characteristic Zipfian distribution of the occurrence of words/tokens in the NMT corpora; which, according to the central thesis of the long-tail theory \cite{feldman_theory}, would lead to memorizations. Similarly, noise in the form of invalid references is an artifact of the scale at which web-based corpora are collected and given that both hallucinations under perturbations and natural hallucinations are widely reported in large-scale NMT systems, our insights should be directly applicable to larger-scale models as well.

We hope that our work serves as a useful step towards a detailed understanding of hallucinations in NMT and in other sequence to sequence models. Among the numerous interesting directions for follow-up work, in future, we would like to explore learning-centric fixes to ameliorate the impact of memorization and corpus-level noise patterns in NMT training.

\label{sec:bibtex}
\bibliography{anthology,custom}
\bibliographystyle{acl_natbib}

\appendix

\section{Appendices}
\label{sec:appendix}
Table \ref{tab:random_examples} presents some random samples from the lower-end of the memorization values. The sentences differ in terms of their syntactic properties (versus Table \ref{tab:mem_examples} in section \ref{ssec:mem_samples}), although more analysis is required to quantitatively ascertain the differences.

\begin{table*}
\centering
\scalebox{0.75}{
\begin{tabular}{l|l|l}
\hline
MV & \textbf{Source}& \textbf{Target}\\ \hline
0.0    & sie haben das gleiche gehirn , und das gleiche eeg . & they have the same brain , and the same eeg .\\ 
0.08    & und wir hatten keine ahnung , wo er war . & it looks like the technological problem is solved .\\ 
0.25    & also , die wiederbesiedlung ist wirklich sehr langsam . & so , recolonization is really very slow .\\
0.30 & zum einen waren beide sehr real . & for one thing , both were very real . \\
0.50    & wir überqueren auf alu-leitern mit angefügten sicherheits-seilen . & we cross on aluminum ladders with safety ropes attached .\\ \hline
\end{tabular}}
\caption{Examples of Samples with Memorization Value (MV) between 0.0 and 0.6 (Below Top-100) from the Training Set as measured using the Memorization Value (MV) Estimator (Algorithm \ref{algo:feldman}) with chrF as the Metric: De-En.}
\label{tab:random_examples}
\vspace{-1.00em}
\end{table*}

\end{document}